# A Hybrid Deep Learning and Model-Checking Framework for Accurate Brain Tumor Detection and Validation

Elhoucine Elfatimi[1*], Lahcen Elfatimi[2] and Hanifa Bouchaneb[3]

[1]Pathology Lab Medicine, school of medicine , University of California, Irvine, CA, USA.

[2,3]Department of Computer Engineering , Polytechnic School of Montreal, QC, Montreal, Canada.

*Corresponding author(s). E-mail(s): eelfatim@uci.edu ;

**Abstract**

Model checking, a formal verification technique, ensures systems meet predefined requirements, playing a crucial role in minimizing errors and enhancing quality during development. This paper introduces a novel hybrid framework integrating model checking with deep learning for brain tumor detection and validation in medical imag- ing. By combining model-checking principles with CNN-based fea- ture extraction and K-FCM clustering for segmentation, the proposed approach enhances the reliability of tumor detection and segmentation. Experimental results highlight the framework's effectiveness, achiev- ing 98% accuracy, 96.15% precision, and 100% recall, demonstrating its potential as a robust tool for advanced medical image analysis.

# 1 Introduction

Model checking is an automatic technique for verifying the correctness prop- erties of safety-critical reactive systems. This method has been successfully applied to find subtle errors in complex systems. Model checking techniques have a wide range of application domains, among which large-scale distributed systems [1–3], signal [4], and medical images analysis [5–8]. The research related to the last topic is still ongoing looking for the perfect (precise, complete, simple) approach for analyzing medical images. The use of model checking is relatively recent, in particular regarding the verification of the analysis of medical images. In this domain, model checking in medical images has shown to be a promising application that can significantly facilitate the work of professionals. What motivates us in this study, considering that model checking is increasingly used in testing to check whether a system model sat- isfies a property, is to take model checking in its usual role to take on more advanced roles in medical image analysis by applying model-checking logic to medical images and detection of tumors in addition to validation of properties through tests or case studies.

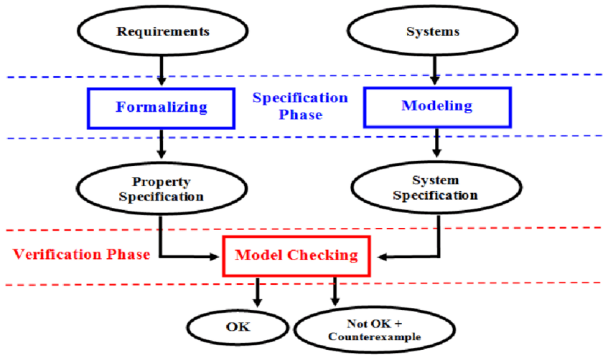

**Fig. 1** Schematic view of model-checking approach

As illustrated in Figure 4, the model-checking approach involves systemat- ically verifying whether specific properties are satisfied across different states within a system. This schematic provides a high-level overview of how model checking is utilized to ensure correctness by exploring all possible states of a system under predefined logical constraints. Such a framework is pivotal in this study as it forms the foundation for integrating model checking with medical image analysis, particularly for detecting and validating tumor regions within brain images.

Brain segmentation, or the process of identifying and separating differ- ent structures or regions within a brain image, has recently gained significant traction in the research area. Segmentation techniques have a variety of appli- cations in neuroimaging, including the analysis of brain scans to identify abnormalities or track the progression of conditions such as brain tumors or neurological disorders. There have been numerous studies on brain segmentation in recent years, and it is considered to be an active and growing area of research. Automatic or semi-automatic segmentation methods, which use algorithms to identify and distinguish different structures in brain images, are often used to improve the accuracy and reliability of brain image analysis. These methods may have a range of applications in the medical field, including diagnosis, treatment planning, and monitoring of treatment response [10, 11]. In [12], the authors suggest a method that involves combining region growing and mathematical morphology to distinguish and separate various structures within the brain.

Segmentation techniques are typically categorized into two groups: generative and discriminative models. Generative models are based on past knowledge of the anatomy of particular brain tissues. In contrast, the models that are discriminative rely on extracting many low-level features in images that can include texture features and local histograms [13]. Generative models use information about the expected appearance and structure of different brain tissues to guide the segmentation process. They may involve creating a statistical model of the appearance of each tissue type and using this model to classify each pixel in the image [13]. Generative models can be effective for segmenting brain images, but they may require a detailed understanding of the appearance and anatomy of the tissues being segmented and may not be as flexible as discriminative models in adapting to variations in image appearance. Discriminative models, on the other hand, use image features such as local histograms or texture patterns to differentiate between different tissue types. These models do not rely on past information about appearance of the tissues, but rather learn to distinguish between different tissues based on the image data itself. Discriminative models may be more flexible and adaptable than generative models, but they may also require a more extensive training dataset and may be more sensitive to variations in image quality or noise [13].

Deep learning, particularly convolutional neural networks (CNNs), has revolutionized the field of medical image analysis by providing state-of-the-art performance in tasks such as segmentation, classification, and feature extraction. CNNs excel at automatically learning hierarchical feature representations directly from raw image data, eliminating the need for manual feature engineering. By leveraging large datasets and powerful computational resources, deep learning models can adapt to diverse imaging modalities and achieve high accuracy in detecting subtle differences in medical images. Despite their success, integrating deep learning with formal verification techniques like model checking remains an underexplored area with significant potential for improving the reliability and interpretability of automated systems.

The fundamental principle of segmentation, which is used in a variety of techniques to take advantage of the differences in intensity and texture among the pixels in an image to identify and distinguish different structures or regions [14]. Segmentation methods may use various techniques to analyze the inten- sity and texture characteristics of the pixels in an image, such as thresholding, clustering, region growing, or machine learning algorithms [14]. These meth- ods aim to separate the pixels into different groups or segments based on their intensity and texture characteristics, with each group corresponding to a dif- ferent structure or region in the image. Segmentation can be a challenging task, particularly in medical images, due to the complexity and variability of the structures being segmented and the presence of noise or other artifacts in the image. However, it is a crucial step in different tasks for image analysis, and various techniques have been developed to improve the accuracy and reli- ability of segmentation methods [14]. Several techniques can also be combined to form hybrid methods. Support vector machines as well as neural networks, are examples of semi-automatic techniques. However, automated segmentation techniques constitute the current gold standard [13–15].

One challenge in developing automated methods for brain tumor segmen- tation is that tumor regions are often only categorized by subtle changes in the brightness of image pixels relative to the surrounding normal tissue [13]. This can make it difficult for automated systems to accurately identify and distinguish tumor regions, as even manual segmentation by experts can exhibit significant variation. This variation may be due to subjective interpretation by the experts or differences in the images of different patients due to the use of different imaging modalities [13]. To address these challenges, automated brain tumor segmentation methods may need to take into account the vari- ability in image appearance and the subjectivity of human interpretation. This may involve using robust image features or machine learning tools to improve the segmentation method's sensitivity and specificity or incorporating expert knowledge or guidance into the segmentation process [13].

This paper introduces a novel framework that integrates model-checking techniques for the verification and validation of brain tumor detection in medical imaging. The key contributions of this work are as follows:

- Development of a new method for checking the detection of brain tumors based on model checking.
- Introduction of a novel logical connector to accurately identify and delineate tumor regions.
- Application of model-checking techniques to enhance the analysis of medical images.
- Demonstration of the pivotal role of model checking in improving accuracy and reliability in medical image analysis.
- Integration of deep learning techniques, specifically CNN-based meth- ods, with model-checking principles to achieve robust and reliable tumor detection.

The remainder of this paper is structured as follows: Section 2 discusses related works, Section 3 outlines the methodology and materials used, Section 4 presents experimental results, and Section 5 concludes with insights and directions for future research.

## 2 Related work

The use of model checking in the context of medical imaging for spatial analysis is a relatively new area of research, as reflected by the small number of pub- lished works in the field. However, the existing studies that have been published are comprehensive in nature. For example, [16, 17] have used machine learning techniques and features of images to validate spatio-temporal models for the detection of tumors. In [18], certain biological processes were focused where model checking was applied, particularly in regards to multi-scale aspects. In contrast, when it comes to fully automated approaches, it is worth noting the significant impact of machine learning techniques that utilize deep learn- ing to model non-linearities in data in order to draw meaningful conclusions [28]. Manually segmenting in vivo images is currently considered the standard practice but it has several disadvantages. The process is time-consuming and requires a high level of skill and attention from the clinician, which makes it costly [14]. Additionally, the subjectivity of manual segmentation can lead to differences in judgment between clinicians and can make it difficult to reproduce, introducing the possibility of human error [14]. Deep learning and machine learning approaches have demonstrated impressive results in tasks such as detection, classification, and pattern recognition. However, the success of these techniques depends on the reliability and quality of the datasets used. Deep learning approaches in particular, require large datasets that capture a wide range of variations. One major advantage of deep learning is its ability to extract features from raw data and make inferences across multiple lay- ers. However, using deep learning in a supervisory framework requires using ground truth labels, which can be subjective and vary significantly between different annotators. According to [14], the intra-expert and inter-expert vari- ability for manually segmenting tumor regions in the brain MRI images can be as high as 35% and 30%, respectively. To address this issue, it is important to use interactive model-checking methods to improve the reproducibility and efficiency of ground truth annotation. Model checking on spatio-temporal data is a relatively new area of research with a limited amount of published litera- ture. However, some notable works in this field include [30], who developed a spatial extension of signal temporal logic for stochastic population models in a discrete representation of space. The framework uses a graph-based approach with weighted, cost-based graphical connections and a single spatial operator similar to the one defined in [20]. [14] extended this approach with the addition of a bounded surrounded operator, which builds upon the surrounded operator described in [21] with additional requirements. [22] alternatively character- ized the bounded surrounded operator as a derivation from a basic operator [21], [23] made significant contributions to the research on model checking as well as spatial logic in medical imaging by developing the theory of models of space utilizing arbitrary graphs, specifically by using a generalized form of topological spaces called 'closure spaces' [24]. This led to the formal defini- tion of spatial logic strictly defined for closure spaces and the development of a model-checking algorithm for it. These works have had a significant impact on this field of research. Model checking has also been applied to the con- touring of nevus images, which are benign skin lesions that can be hard to identify from melanoma, deadly skin cancer. For effective disease management and treatment, early detection and diagnosis are crucial. In [25], the authors used texture similarity operators and spatial logic operators to tackle the chal- lenges of contouring 2D nevus images, which can vary in texture, size, color, and shape, and may contain extraneous elements. They tested the effective- ness of their technique on a public dataset. While model checking has shown comparable quality to the current gold standard for brain tumor segmentation [14], [26, 27], additional challenges such as optical effects and variance within types of lesions make nevus segmentation and model checking is more com- plex. Often model checkers utilize specifications that are high level and written in logic language so as to model spatial properties in order to identify specific patterns which in turn are key in the identification of structures such as in [25]. In [25], the authors investigate the feasibility of a technique that is based on spatial logic for closure spaces for analysis of nevi images from a public dataset. Authors use spatial model checking techniques to analyze nevi images from a public dataset. They show that these techniques can effectively identify spatial patterns in the images, even with the presence of inhomogeneity and extraneous elements. The results of the analysis are compared to the manually- annotated ground truth data from the dataset. The authors suggest that the use of logic-based approaches and efficient implementation of spatial model- checking algorithms contribute to the success of the analysis in this context. The dataset is provided by Skin Lesion Analysis toward Melanoma Detection challenge 2016 [28]. The inter and intra variations in the nexus, as reported in [14] pose a great challenge in segmentation. Thus, the approach is to dis- tinguish between nevus tissue and skin tissue. The authors use a statistical texture analysis operator to approximate a nevus. The algorithm takes local histograms into account to determine between background pixels and target pixels. The texture operator used for this serves as a good first approximation but requires adjustments, such as in the form of the use of derived operators with metrics like similarity indexes. The spatial model checking techniques can be used with the spatial model checking tool VoxlogicA developed in [29] to efficiently segment nevi. Inspired by closure spaces, [14] presented a novel segmentation method to combine spatial operators and domain-based opera- tors. The authors explore a semi-automatic contouring approach in [29] with a tool VoxLogicA that takes advantage of the library of computational imag- ing algorithms alongside distinct combinations of the declarative specification to deliver optimized execution. This approach has been shown to achieve high accuracy and also has the advantage of being easily explainable and replica- ble. Spatio-temporal model checking involves describing spatial characteristics by leveraging logical language to identify important structures and patterns. The difficulty in accurately identifying tumors arises from the fact that they are only distinguishable from normal tissue based on slight variations in pixel intensity in grayscale images. Additionally, there is significant subjectivity in the manual segmentation of ground truth images by experts, which can vary greatly and make it difficult to isolate tumors from normal tissue. This is fur- ther complicated by variations in image quality among different MRI scanners [29]. For demonstration, authors use the publicly available BraTS 2017 dataset [13]. Authors build on the image query language that was proposed by [15] which in turn was inspired by spatial logic for closure spaces [23]. Authors in

[29] derive their kernel from [23] for their framework. This work [29] is closely related to [13] with respect to the distance-based operator. For digital image analysis, a statistical similarity operator is employed to compare the similarity of a particular area in an image with a specified region. This is done by calcu- lating histograms of both areas and determining the cross-correlation between them. This operator allows for evaluating how similar the area around a point is to a particular region based on statistical measures. The percentile operator is another tool used in digital image analysis. It takes a numeric-value image and its binary mask as input and produces an output image that shows the percentile rank of the intensity of each point in relation to the intensity of all the voxels in the population. This operator enables the use of the same seg- mentation specifications on images with different intensity distributions and avoids the need to use absolute values in constraints on intensity. VoxLogicA is a tool designed for multidimensional images and their spatial analysis. It operates by interpreting a specification written in the image query language and producing a set of multidimensional images that represent the evaluation of the user-specified expressions. In the case of medical images, boolean values are used for logical operators, allowing the regions of interest to be overlaid on the original images for clearer visualization. Non-logical operators, on the other hand, produce number-valued images. The pipeline for using VoxLogicA is straightforward, as it simply requires the input of a specification written in the appropriate language. The tool was by [29] for evaluation for VoxLogicA for the segmentation of Glioblastoma as well as on the BraTS 2017 dataset. A variant of the method described in a previous study [15] was used to compare the performance of VoxLogicA with another model called topochecker. In order to perform the analysis, topochecker takes 52 seconds for two-dimensional images as compared to the 750 milliseconds that VoxLogicA takes. For three-dimensional images, topochecker takes 30 minutes compared to only 15 seconds that VoxLogicA takes. The significant improvement in performance mentioned earlier was achieved through the use of a specialized imaging library and the implementation of new algorithms, such as the statistical similarity operator, as well as parallel execution and other optimization techniques

In [27], the main focus is on using model checking in imaging to create rep- resentations of human body parts for possible medical intervention planning as well as clinical analysis, with a particular emphasis on accurate contouring tissues and organs in the healthy brain. Automated software tools that are commonly used for this purpose are often highly specialized and lack flexibility and transparency, and may not always produce accurate results. Deep learning- based approaches have recently gained popularity for medical image analysis, but they require large, accurately labeled datasets that are not always avail- able and can be prone to variability in labeling by experts. The framework for the spatial logic utilized by [27] involves modeling an image as an adjacency space. This means that only the pixels that share an edge are counted as adja- cent. The adjacency space is a subclass of closure spaces. Closure spaces can be strengthened by means of a distance metric, thus, leading to distance clo- sure spaces. In the study by [27], several new operators were introduced for use in the spatial model checking framework. The "touch" the operator ensures that starting and ending points meet certain criteria and all intermediate points between them. The "grow" operator expands a region based on speci- fied criteria. The "filter" operator performs a smoothing operation by taking into account a specified radius around each point. The "statistical similarity" operator compares the histograms of two regions to identify tissues with sim- ilar textural characteristics. This operator is based on cross-correlation and is invariant to rotation, making it particularly suitable for use in spatial logic for medical applications. The performance of these operators was demonstrated using a benchmark checkerboard pattern, and the authors also applied them to the segmentation of simulated brain images. These images are particularly useful for testing methods that aim to identify different types of brain tissue, such as gray matter and white matter.

## 3 Method and Material

This section discusses the application of model checking for tumor detection, including the methodology, dataset, and tools used

### 3.1 Application of the Model-Checking for Tumor Detection

Model-checking techniques provide an opportunity to reduce testing costs and enhance confidence in underdeveloped systems. Furthermore, model checking minimizes the risk of introducing errors during the early stages of system development [30]. Given this role of model checking in validation and error reduction, we address one of the primary challenges in utilizing model checking: studying and validating the satisfaction of properties for each point $x \in X$ within the Γ model.

To apply model-checking logic to medical images and determine its effec- tiveness in a brain tumor detection study, several essential characteristics must be defined:

1. Define the input model Γ.
2. Define the properties to be checked.
3. Determine the set of states, then specify the tumor region.
4. Check the ownership satisfaction at each point.

Our first objective is to define the input model and determine the set of states (*the set of points*) before detecting the tumor region. Subsequently, we test and study each case to determine if the tumor is detected correctly, followed by a comparison of results.

**Syntax**

The syntax of the logic used in this paper is as follows: If $\phi$ and $\psi$ are CTL formulae, then so are:

$$\neg\phi, \phi \wedge \psi, \phi \vee \psi$$

**EX$_\psi$**: holds in some next state
**EF$_\psi$**: along some path, $\psi$ holds in a future state
**EG$_\psi$**: along some path, $\psi$ holds in every state.

To determine the entry model, we rely on the following definition:

**Definition 1** Model $\Gamma$ is a 4-tuple $(C, V, A, T)$ where:

- $C$: is the finite set of states;
- $V$: is the finite set of variables;
- $A$: is a set of actions;
- $T$: is a finite set of transitions.

### 3.1.1 Properties satisfaction verification

The method we used in our work is based on creating some derived opera- tors instrumental in analyzing medical images to verify the satisfaction of the properties $\varphi 1$ and $\varphi 2$ in different cases, which helps us check tumor detection.

**Definition 2** We define some operators that we used to analyze medical images to verify the properties $\varphi 1$ and $\varphi 2$ in different cases.

$$border, x| = \varphi 2 \qquad (1)$$
$$background, x| = connect(\varphi 1, \varphi 2); \qquad (2)$$
$$brain, x| = connect(\varphi 1, \varphi 2) \wedge \varphi 2; \qquad (3)$$
$$\Gamma, x| = str(d', \varphi 2) \qquad (4)$$
$$\Gamma, x| = increase(\varphi 1, \varphi 2); \qquad (5)$$

- border, x | =$\varphi 2$ is true on voxels that lie on the boundary of the image.
- background, x| = $connect(\varphi 1, \varphi 2)$; defines the background, i.e. all voxels that are part of an area that touches the edge of the image
- brain,x | =!$connect(\varphi 1, \varphi 2) \wedge \varphi 2$: return points that are not satisfy the background and $\varphi 2$.
- The formula connect($\varphi 1, \varphi 2$) is satisfied by the voxels which satisfy $\varphi 1$ and not $\varphi 2$ and which have a distance from a state $\varphi 2$ (Fig 3.A).

- The formula Str(d', φ2 ) is at least a state distance that does not satisfy φ1, and is satisfied by states that are less than d'(Fig 3.B).
- The formula increase (φ1, φ2) is satisfied by states which satisfy φ2 and satisfy φ1, and which can reach a satisfactory state φ1 (Fig 3.C).

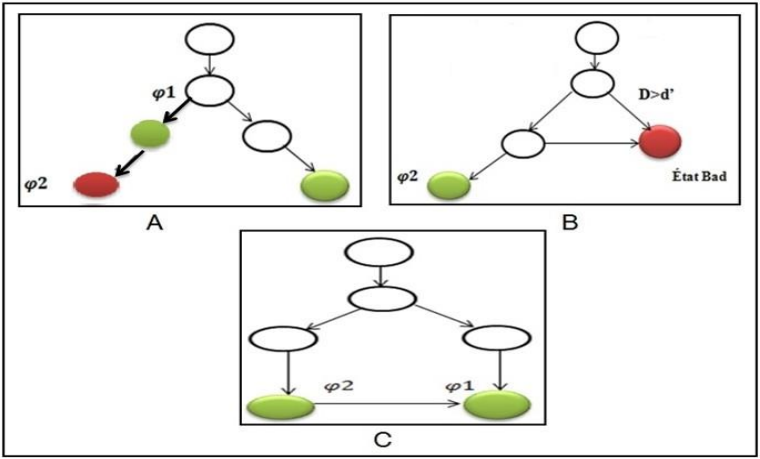

**Fig. 2** Logical operators for medical image analysis

In conclusion, the specification of logical operators helps us verify tumor detection and strengthens the role of model checking in medical image analysis.

## 3.2 Methodology

This work will focus on analyzing the effectiveness of the model-checking algo- rithm in quantifying the quality of brain tumor detection using the CNN algorithm. In addition to determining whether the tumor detection was done correctly, the study of whether the model satisfies the properties. Fig 3 is shown the methodology we follow in our work.

The methodology of this work is mainly concerned with performing a model check based on the implemented segmentation of the input image and detecting the tumor using the CNN algorithm. The essential steps in this methodology are as follows.

1. **Input Image:** Input images are implemented using the medical Brain MRI images dataset.
2. **Preprocessing:** The combined input images are subject to pre-processing. In the preprocessing step, we can perform: image enhancement and image resize conversion, as well as grayscale conversion, is performed.
3. **Segmentation:** FCM Clustering is used to segment the data in this step. Then, a cluster is chosen from among the others that are appropriate.

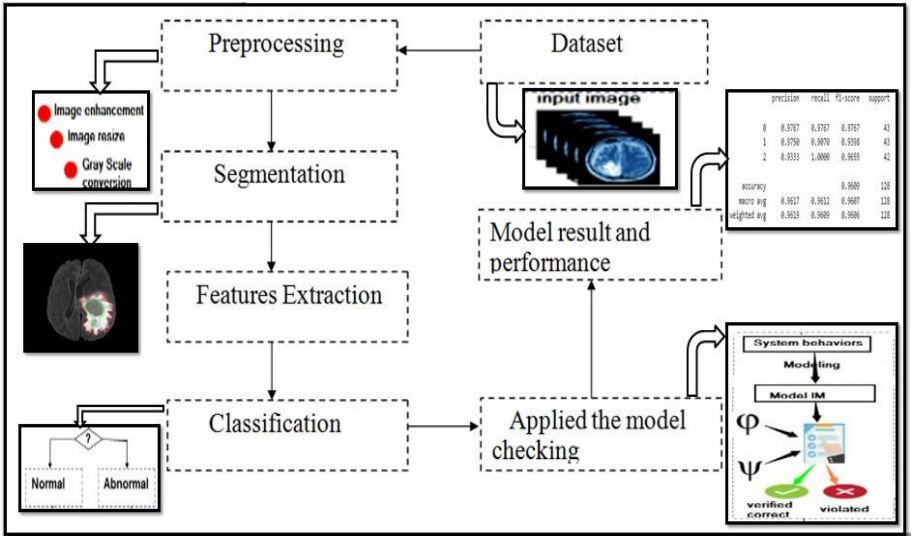

**Fig. 3** Overall block diagram of proposed methodology

4. **Feature Extraction:** Feature extraction has been performed using Pattern and texture.
5. **Classification and detection:** The CNN algorithm is used to detect a tumor.
6. **Applied the model checking :** Analyzing the effectiveness of the model checking algorithm in quantifying the quality of brain tumor detection using the CNN algorithm. In addition to determining whether the tumor detection was done correctly.
7. **Model result and performance :** In this section, the performance metrics like Accuracy, Sensitivity, and Specificity will be estimated. The performance of Model Checking and CNN will be compared by means of the performance parameters.

## 3.3  Data set

This study used a publicly available dataset selected from Kaggle. It contains two categories: a tumor category with several cases of tumors that differ in location, size, and speed of spread, and a healthy category that does not include tumors. The tumor category comprises 1550 images, and the non-tumor cate- gory comprises 980 images. This data set was generated in 2019 by Anshu S et al. More than 38,440 people downloaded it for its effectiveness in evaluat- ing the quality of the algorithms to be studied. The image below shows some samples from the data set.

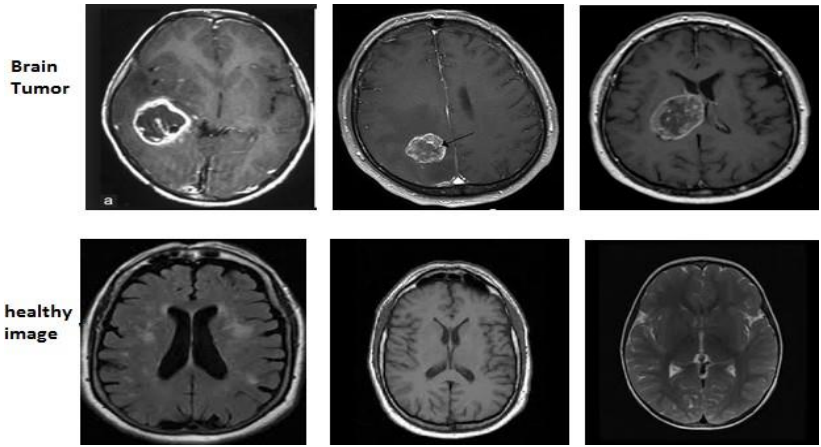

**Fig. 4** Brain tumor and healthy image.

## 3.4 Implementation Tools and Frameworks

This study utilized a combination of tools and technologies to implement and evaluate the proposed approach. These tools were selected for their effectiveness in handling different stages of the workflow:

- **C**: The C programming language was used to implement the core logi- cal operators and the model-checking algorithms. Its high performance and efficient memory management made it ideal for executing computationally intensive validation tasks.
- **MATLAB**: MATLAB was utilized for preprocessing the medical images, including tasks such as resizing, filtering, and grayscale conversion. Addi- tionally, it was employed for clustering and feature extraction, leveraging its extensive libraries for image processing and analysis.
- **Python with TensorFlow/Keras**: Python, combined with the Tensor- Flow/Keras deep learning framework, was used to implement the Convolu- tional Neural Network (CNN) for classification tasks. The CNN was designed to distinguish between tumor and non-tumor images, utilizing its ability to automatically extract and learn features from the medical imaging dataset. This facilitated accurate and efficient tumor detection.

The integration of these tools provided a comprehensive environment for implementing the proposed hybrid approach, combining model-checking prin- ciples with deep learning-based classification to ensure high accuracy and reliability in tumor detection.

The combination of these tools provided a balance of performance and flexibility, enabling seamless integration of model-checking principles with advanced image analysis techniques.

# 4 RESULTS

The most important result of this work is the use of model checking to study tumor detection. This approach extends the traditional roles of model checking to advanced applications in medical imaging by developing a methodology that determines whether tumor region detection is correct, while also studying the satisfaction of possible properties and extracting detailed information.

The proposed models included a dataset of 2530 images. Initially, input images with dimensions of 1427 (rows) and 1275 (columns) undergo a pre- processing stage. During this step, the images are resized to dimensions of 256 (rows) by 256 (columns), converted to grayscale, and processed with a filter. The filter parameters include a filter size of 3 × 3 and a smoothing level of 0.5, as illustrated in Figure 8.

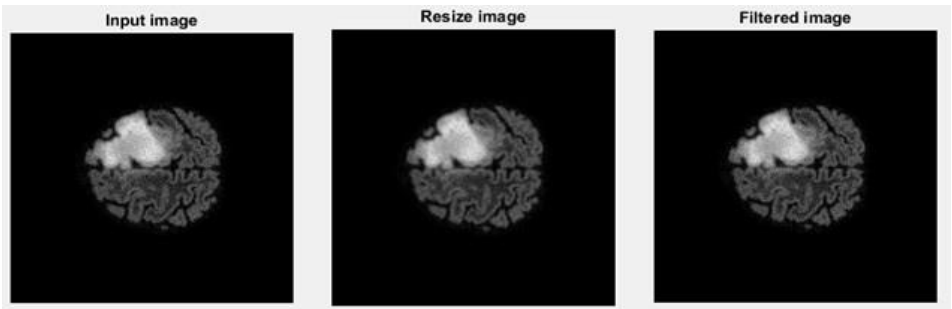

**Fig. 5** Preprocessing steps: resizing, grayscale transformation, and filtering.

Tumor segmentation is performed using K-FCM clustering. This clustering process isolates the region of interest from the background. Following cluster- ing, color-based segmentation is applied to further delineate the tumor area. The number of clusters specified in the implementation is 4, resulting in the image being partitioned into four distinct regions.

**Algorithm 1** Segmentation

**Require:** global *Filt Filt*1, global *SEG RES*
1:  **procedure** TUMOR SEGMENTATION()
2:      numofclusters = 3;
3:      Filt1(:,:,1) = Filt;
4:      Filt1(:,:,2) = Filt;
5:      Filt1(:,:,3) = Filt;
6:      [insegment,segment,center ] = KFCM( double(Filt1), nanoclusters );
7:      axes(handles.axes4) ;
8:      imshow(uint8(segment));
9:   Segmented2 = zeros(size(segment,1),size(segment,2));   10: Segmented3 = zeros(size(segment,1),size(segment,2));   11: Segmented4 = zeros(size(segment,1),size(segment,2));   12:   **for** ii = 1:size(segment,1) **do**
13:          **for** jj = 1:size(segment,2); **do**
14:              **if** segment(ii,jj) == 25 **then**
15:                  Segmented1(ii,jj,1) = Filt1(ii,jj,1);
16:                  Segmented1(ii,jj,2) = Filt1(ii,jj,2);
17:                  Segmented1(ii,jj,3) = Filt1(ii,jj,3);
18:                  **if** $segment(ii, jj)$ == 50 **then**
19:                      Segmented2(ii,jj,1) = Filt1(ii,jj,1);
20:                      Segmented2(ii,jj,2) = Filt1(ii,jj,2);
21:                      Segmented2(ii,jj,3) = Filt1(ii,jj,3);
22:                  **else** $segment(ii, jj)$ == 100
23:                      Segmented3(ii,jj,1) = Filt1(ii,jj,1);
24:                      Segmented3(ii,jj,2) = Filt1(ii,jj,2);
25:                      Segmented3(ii,jj,3) = Filt1(ii,jj,3);
26:                  end
27:                  Else $segment(ii, jj)$ == 150
28:                      Segmented4(ii,jj,1) = Filt1(ii,jj,1);
29:                      Segmented4(ii,jj,2) = Filt1(ii,jj,2);
30:                      Segmented4(ii,jj,3) = Filt1(ii,jj,3);
31:              endfor
         endfor
32:
33:

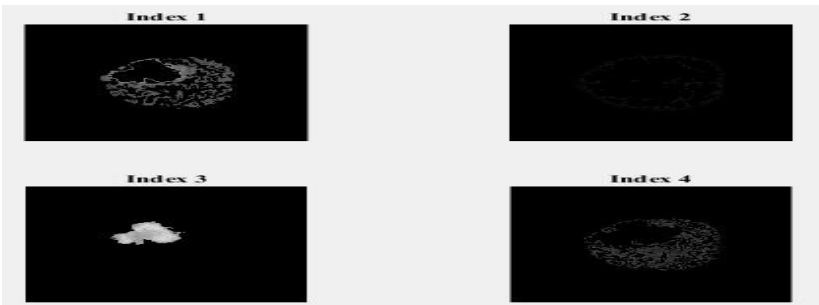

**Fig. 6** Select the appropriate region

Algorithm 2 is a switch that enables us to choose the case that is closest to the tumor and the most obvious from among the four images, and among that four regions, we have to select the appropriate region, then that region will be the segmented image.

---

**Algorithm 2** select the applicable Tumor image **Require:**
INPT = input('Enter the Index num :: '); 1: **procedure**
SELECT REGION()
  2:     **switch INPT**
  3:     **case 1 :**
    SEG = Segmented1;
  4:     **case 2**
    SEG = Segmented2;
  5:     **case 3**
    SEG = Segmented3;
  6:     **case 4**
    SEG = Segmented4; end
  7:  **end procedure**
        **Output** : selcted
        image =0

---

The result is a trace as shown below, where the most obvious case number is entered for segmentation.

Iteration count = 1, obj. fcn = 130738752.442221 Iteration count = 2, obj. fcn = 101233315.647825 Iteration count = 3, obj. fcn = 101028357.292125 Iteration count = 4, obj. fcn = 98824526.226355 Iteration count = 5, obj. fcn = 87246159.498379 Iteration count = 6, obj. fcn = 64942715.789979 Iteration count = 7, obj. fcn = 50294485.599759 Iteration count = 8, obj. fcn = 43571128.971012 Iteration count = 9, obj. fcn = 34130654.776338 Iteration count = 10, obj. fcn = 25372817.199223 Iteration count = 11, obj. fcn = 21506791.474057 Iteration count = 12, obj. fcn = 20363031.608788 Iteration count = 13, obj. fcn = 20094601.506415 Iteration count = 14, obj. fcn = 20039248.495941 Iteration count = 15, obj. fcn = 20028426.511818 Iteration count = 16, obj. fcn = 20026348.823408 Iteration count = 17, obj. fcn = 20025952.298998 Iteration count = 18, obj. fcn = 20025876.786704 Iteration count = 19, obj. fcn = 20025862.419908 Iteration count = 20, obj. fcn = 20025859.687728

Iteration count = 21, obj. fcn = 20025859.168260 Iteration
count = 22, obj. fcn = 20025859.069505

Enter the Index num ::     3

For Figure 4 above, index 3 is applicable. CNN is subsequently used to classify index 3 as either tumor or non-tumor. If a tumor is detected, a message will be displayed, as shown in Algorithm 3.

---

**Algorithm 3** Check for tumor

---

**Require:** global $Features_R result$, global $TestfeaLabel$, load $Trainfea1$

1: **procedure** MESSAGE ()
2:     **for** ijk = 1:size(Trainfea1,1) **do**
3:         temp = Trainfea1(ijk,:);
4:         Dist val(ijk) = mean(temp - Testfea);
5:     **end for**
6:     **if** result(1) == 1 **then**
7:         msgbox(' Normal');
8: **else** result(1) == 2
9:     msgbox(' Abnormal');
10:

---

Then the validity of detection is checked using model checking based on property satisfaction studies. If the model satisfies the properties, the tumor is confirmed, as shown in Algorithm 4. Here, we input $\varphi$, and a finite model $\Gamma$ to determine the set of points $X$ satisfying $\varphi$. Where we enter the input $\varphi$, and a finite model $\Gamma$ the objectif is checking the set of points X satisfying $\varphi$

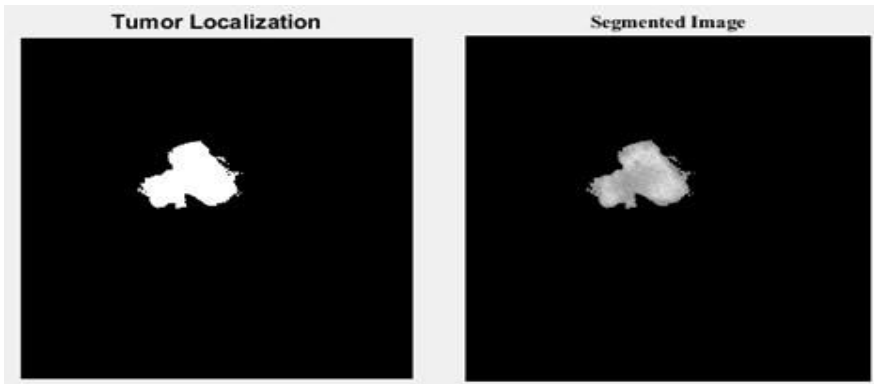

**Fig. 7** Tumor localization and Segmented image

**Algorithm 4** Validation of Tumor Segmentation

**Require:** A formula $\varphi 1$, $\varphi 2$ and a finite model Γ, img = "Dataset"
1: **procedure** TUMOR DET SAT($\varphi 1, \varphi 2$, Γ)
2:    Input img
3:    image enhancement
4:    image resize
5:    grayscale conversion
6:    border=$\varphi 2$
7:    b = connect($\varphi 1$, 0,9, $\varphi 2$);
8:    brain=connect( $\varphi 1$, $\varphi 2$) ∧$\varphi 2$: return the voxels that do not satisfy the background
9:    Update reference pixel value if needed.
10:    initial tumor location 11: Final tumor location 12: **end procedure**
   **Output**: Final tumor segmentation
      =0

## 4.1 Performance evaluations

To evaluate the effectiveness of the classification system, we employed the confusion matrix, a comprehensive tool for assessing classification accuracy. The confusion matrix provides a detailed breakdown of the system's performance by categorizing predictions into True Positives (TP), True Negatives (TN), False Positives (FP), and False Negatives (FN). This approach allows for immediate identification of the classification system's accuracy and reliability.

Tumor segmentation outcomes were assessed using key performance metrics derived from the confusion matrix:

$$Precision = \frac{TP}{TP + FP} \quad (6)$$

$$Recall = \frac{TP}{TP + FN} \quad (7)$$

$$Accuracy = \frac{TP + TN}{TP + TN + FP + FN} \quad (8)$$

$$F1 = \frac{2 * precision * recall}{precision + recall} \quad (9)$$

Using the confusion matrix and calculating performance metrics for both the current and proposed methods, we achieved significant results, with the accuracy reaching 98%, surpassing previous works that did not exceed 97%. This is demonstrated in Table 1, where the integration of model checking with machine learning approaches resulted in enhanced performance. These findings underscore the pivotal role of model checking in advancing the analysis of medical images.

**Table 1** Performance parameters of the existing and proposed methods.

| Technique | Accuracy | Precision | Recall |
|---|---|---|---|
| **CNN [31]** | 96.17% | 96.17% | 96.12% |
| **Markov model(DTMC) [32]** | 97.65% | 71.65% | 99.87% |
| **Cellular Automata [33]** | 93 % | 95% | 90% |
| **Model Checking** | 98% | 96.15% | 100% |

To further illustrate the results, Figure 8 compares the proposed method- ology with prior approaches. This comparison relies on the same dataset while employing different techniques, such as the Markov model, cellular automata, and CNN. The current work, which integrates machine learning approaches with model checking, demonstrates a notable reduction in errors and an improvement in the quality of tumor detection.

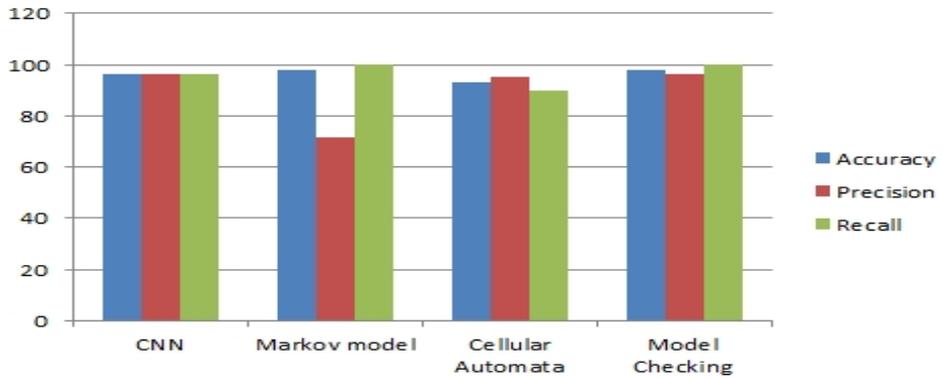

**Fig. 8** Comparison between existing and proposed methods

## 5  CONCLUSION

This work demonstrated the integration of model checking with deep learning techniques for tumor detection, achieving a significant accuracy of 98%. The results highlight the transformative potential of model checking in medical image analysis, enabling robust validation and enhancing the quality of tumor segmentation. By combining model checking with methodologies such as deep learning, the Markov model, and cellular automata, this approach not only reduced errors but also improved diagnostic precision.

Looking ahead, future research will focus on developing more advanced methodologies for tumor segmentation, particularly for tumors located in diverse anatomical regions. This includes creating a specialized specification language for tumor segmentation and treating medical images as detailed mod- els for analysis. Furthermore, designing algorithms to identify similar tumor regions using model checking techniques will refine detection capabilities and promote greater precision in medical imaging.

The success of this study underscores the importance of integrating formal verification methods like model checking with machine learning, paving the way for more reliable and accurate diagnostic tools in the field of medical imaging.

## Declarations

- Funding: This research received no external funding.
- Conflict of interest: The authors declare no conflict of interest.
- Informed Consent Statement: Not applicable.
- Availability of data and materials: The publicly accessible data sets used in this work are available here: https://www.kaggle.com/datasets/navoneel/ brain-mri-images-for-brain-tumor-detection.